\documentclass[journal]{IEEEtran}
\usepackage{amsmath,amssymb,amsfonts}
\usepackage{latexsym}
\usepackage{algorithmic}
\usepackage{graphicx}
\usepackage{multirow}
\usepackage{multicol}
\usepackage{bbm}
\usepackage{bm}

\usepackage{cite}
\usepackage{times}
\usepackage{soul}
\usepackage{url}
\usepackage[hidelinks]{hyperref}
\usepackage[utf8]{inputenc}
\usepackage[small]{caption}
\usepackage{booktabs}
\usepackage{algorithm}
\usepackage[switch]{lineno}
\usepackage[table,xcdraw]{xcolor}
\usepackage{array}
\usepackage{lineno}
\ifCLASSINFOpdf
\else
\fi
\begin{document}
%
\title{PSGR: Pixel-wise Sparse Graph Reasoning for COVID-19 Pneumonia Segmentation in CT Images}
%
%
%

\author{Haozhe Jia,~\IEEEmembership{Student Member,~IEEE,}
        Haoteng Tang,~\IEEEmembership{Student Member,~IEEE,}
        Guixiang Ma, ~\IEEEmembership{Member, ~IEEE}
        Weidong Cai,~\IEEEmembership{Member,~IEEE,}
        Heng Huang,~\IEEEmembership{Member,~IEEE,}
        Liang Zhan,~\IEEEmembership{Member,~IEEE,}
        Yong Xia,~\IEEEmembership{Member,~IEEE}
\thanks{H. Jia and H. Tang contributed equally to this work. Corresponding authors: L. Zhan and Y. Xia}
\thanks{H. Jia and Y. Xia are with the National Engineering Laboratory for Integrated Aero-Space-Ground-Ocean Big Data Application Technology, School of Computer Science and Engineering, Northwestern Polytechnical University, Xi’an 710072, China (e-mail: haozhejia@mail.nwpu.edu.cn; yxia@nwpu.edu.cn).}
\thanks{G. Ma is with the Intel Labs, 2111 NE 25th Ave, Hillsboro, OR, 97124, USA (email: guixiang.ma@intel.com)}
\thanks{W. Cai is with the School of Computer Science, The University of Sydney, Sydney, NSW 2006, Australia (e-mail: tom.cai@sydney.edu.au).}
\thanks{H. Jia, H. Tang, H. Huang, and L. Zhan are with the Department of Electrical and Computer Engineering, University of Pittsburgh, Pittsburgh, PA 15261 USA; H. Huang is also with the JD Finance America Corporation, Mountain View, California, CA 94043, USA (email: haoteng.tang@pitt.edu; heng.huang@pitt.edu; liang.zhan@pitt.edu)}
}

\maketitle

\begin{abstract}
Automated and accurate segmentation of the infected regions in computed tomography (CT) images is critical for the prediction of the pathological stage and treatment response of COVID-19.
Several deep convolutional neural networks (DCNNs) have been designed for this task, whose performance, however, tends to be suppressed by their limited local receptive fields and insufficient global reasoning ability.
In this paper, we propose a pixel-wise sparse graph reasoning (PSGR) module and insert it into a segmentation network to enhance the modeling of long-range dependencies for COVID-19 infected region segmentation in CT images.
In the PSGR module, a graph is first constructed by projecting each pixel on a node based on the features produced by the segmentation backbone, and then converted into a sparsely-connected graph by keeping only $K$ strongest connections to each uncertain pixel. The long-range information reasoning is performed on the sparsely-connected graph to generate enhanced features.
The advantages of this module are two-fold: (1) the pixel-wise mapping strategy not only avoids imprecise pixel-to-node projections but also preserves the inherent information of each pixel for global reasoning; and (2) the sparsely-connected graph construction results in effective information retrieval and reduction of the noise propagation.
The proposed solution has been evaluated against four widely-used segmentation models on three public datasets. The results show that the segmentation model equipped with our PSGR module can effectively segment COVID-19 infected regions in CT images, outperforming all other competing models.
\end{abstract}


%
\IEEEpeerreviewmaketitle

\section{Introduction}
%
%
%
%
\label{sec:introduction}
\IEEEPARstart{T}{he} pandemic of COVID-19 has become one of the most severe global health crises in human history, leading to enormous loss of population and prosperity \cite{wang2020novel}.
Although it has been well recognized as the gold standard for COVID-19 screening, the reverse transcription polymerase chain reaction (RT-PCR) requires a huge amount of human resources and medical equipment and is limited by its high false negative rate \cite{ai2020correlation,fang2020sensitivity}. 
Thanks to the development of the computer-aided systems, a variety of 
medical imaging techniques (\textit{e.g.}, X-rays, computed tomography (CT), and magnetic resonance imaging (MRI)) have been integrated into an uniform platform to detect (\textit{e.g.}, with computer-aided detection (CADe)) and diagnose (\textit{e.g.}, with computer-aided diagnosis (CADx)) the disease \cite{huang2011pacs}.
Among them, CT has been broadly utilized as an assistance to RT-PCR due to its superior imaging quality and 3-dimensional view of the lungs  \cite{ng2020imaging,ai2020correlation,fang2020sensitivity,song2021augmented,zhang2020viral,liu2021special}.

Beyond screening COVID-19 cases, effective detection and segmentation of COVID-19 infection using CT can benefit the prediction of the pathological stage, development, and treatment response of the disease.
Currently, the segmentation is usually conducted by radiologists through visual inspections, which is time consuming, professional-skill intensive, and not applicable to large-scale screening.
The success of deep convolutional neural networks (DCNNs) in many image segmentation applications has prompted investigators to apply DCNNs to this task \cite{xu2020deep,fan2020inf,amyar2020multi,qiu2021miniseg,zheng2020deep,wu2021jcs}.
However, despite several attempts, the segmentation of COVID-19 infection remains challenging due to the speciality of pathological patterns of COVID-19 in CT images, including the fact that infected regions usually (1) vary in shape, size, and location, (2) appear to be visually similar to some surroundings tissues, and (3) disperse wildly within the lung cavity. We believe that the segmentation performance of current DCNN-based solutions tends to be suppressed by their limited local receptive fields and insufficient global reasoning ability.

In recent years, the self-attention mechanism \cite{vaswani2017attention} has been introduced to DCNNs to enhance the long-range dependency of spatial contextual information for semantic segmentation \cite{wang2018non,zhao2018psanet,fu2019dual,huang2019ccnet}. 
Although being able to capture, to some extent, the non-local long-range information, these non-local methods usually construct excessive correlations among all pixels, which may introduce redundant information and suppress the discriminatory power of image features. 
Moreover, the global reasoning ability of these models is still limited, since the interactive information is merely delivered and aggregated at the image level.

Recently, graph neural networks (GNNs) have enjoyed increasing success and advanced ever more powerful in semantic image segmentation, showing great potentials in enhancing DCNNs with the global reasoning ability.
The GNNs are designed to embed the graph node based on the information propagation mechanism, where any node in a graph can gain the information across all nodes from the whole graph \cite{velivckovic2017graph,hamilton2017inductive,scarselli2008graph,gilmer2017neural,battaglia2018relational,defferrard2016convolutional,tang2021commpool}.
In general, GNNs include three operations: message propagation, information aggregation, and feature transformation \cite{ying2019gnnexplainer}. 
The information propagation manner of GNNs can break through the restriction of the local receptive field of the traditional convolution filter and enables the long-range dependency reasoning based on the global feature map.

When incorporating a GNN into a DCNN-based segmentation model, a crucial step is to construct a projection that maps DCNN-generated features to the graph space. 
Usually, a cluster of pixels is identified based on feature similarity and is directly projected onto a graph node \cite{li2018beyond,chen2019graph}. 
This projection scheme requires a predetermined number of nodes, which may not be suitable for all cases.
Alternatively, an image can be partitioned into regions based on its structural information or pseudo landmarks, and each region is then projected onto a node \cite{liu2020cross}. 
This solution relies highly on prior knowledge for image partitioning and has poor generalizability.
Particularly, the infected regions in COVID-19 CT images are usually small, disperse, and morphologically diverse, whereas normal regions are usually large \cite{fan2020inf}.
When using the cluster- or region-based method to convert a COVID-19 CT image into a graph, a huge number of normal pixels might be projected onto one node, which contains too much diverse information that could suppress other nodes, especially those which represent small infected regions, in the graph reasoning stage.
Moreover, due to the inaccuracy of pixel clustering and image partitioning, both projection schemes may assign a few normal pixels to a node that represents an infected region and vice versa. Such inaccuracy may disturb subsequent pixel classification. 

To address these drawbacks, an intuitive solution is to pixel-wisely map each pixel to a node, which, similar to those non-local methods, would build a densely connected graph. 
However, such a solution may be intractable due to its extremely high computational and spatial complexity. 
Moreover, it is not reasonable and necessary for each node to gain effective information from all other nodes, since extra noise may be introduced during this process \cite{Hou2020Measuring}.

In this paper, we propose a pixel-wise sparse graph reasoning (PSGR) module and incorporate it into a backbone network for the segmentation of COVID-19 infection in chest CT images.
Specifically, the workflow of PSGR module consists of three steps. First, a densely-connected graph is constructed by projecting each pixel onto a node based on the features generated by the segmentation backbone. 
Second, the graph is converted into a sparsely-connected one by keeping only $K$ strongest connections to each uncertain node (pixel), where the strength of a connection is measured by the similarity of its two nodes in the feature space and the uncertainty of each node is determined by an additional coarse segmentation branch.
Third, the long-range information reasoning is performed on the sparsely-connected graph and the enhanced features are generated and fed to the segmentation backbone. 
The proposed solution has been evaluated against widely-used segmentation models on three public CT datasets for both bi-class and multi-class COVID-19 infection segmentation.

The main contributions are summarized as follows:
\begin{itemize}
    \item An intuitive pixel-wise mapping strategy is used to construct the graph, leading to two benefits: (1) avoiding imprecise pixel-to-node projections and (2) preserving the inherent information of each pixel for reasoning long-range contextual information.
    \item An edge pruning method is proposed to convert the graph into a sparsely-connected one, resulting in effective information retrieval and reduction of the noise propagation among nodes.
    \item The results on three datasets show that the backbone network integrated with the proposed PSGR module is superior to all competing models in the segmentation of COVID-19 infected regions in CT images, and the ablation study also demonstrates the effectiveness of our PSGR module.
\end{itemize}

\section{Related Work}
\label{sec:related work}
\subsection{Segmentation of COVID-19 Infection in CT Images}
With the successful application of DCNNs to medical image segmentation, various DCNNs have been proposed to segment COVID-19 infected regions in chest CT images.
Xu \textit{et al.} \cite{xu2020deep} introduced a region proposal network to a residual-inception V-Net for the segmentation of candidate infected regions in CT images.
Fan \textit{et al.} \cite{fan2020inf} developed a novel COVID-19 infection segmentation network called Inf-Net, which utilizes the reverse attention and edge-attention to improve the performance and also employs the semi-supervised learning to alleviate the shortage of high-quality annotations.
Amyar \textit{et al.} \cite{amyar2020multi} proposed a multitask deep learning model to jointly identify COVID-19 patients and segment COVID-19 lesions using chest CT.
This model not only leverages useful information contained in related tasks to improve both segmentation and classification, but also reduces the impacts caused by the small dataset.
Qiu \textit{et al.} \cite{qiu2021miniseg} proposed a lightweight deep learning model called MiniSeg, which reduces the computational cost of training and can segment COVID-19 CT images efficiently.
However, none of these models attempt to explore and utilize the long-range dependencies, which may overlook the rich image contextual information, to improve the performance of the semantic segmentation. 
In this work, we incorporated GNN into a DCNN-based segmentation model to enhance the modeling of long-range dependencies, which plays a pivotal role in improving the accuracy of COVID-19 infection segmentation.

\subsection{Global Contextual Information Learning}
Constrained by the local receptive field of convolutional operations, DCNN-based segmentation models tend to have a limited ability to capture global contextual information. 
To address this issue, \cite{zhao2017pyramid,chen2018encoder,chen2017deeplab,chen2017rethinking} utilized dilated convolutions and pyramid pooling to enlarge the receptive field of DCNNs and showed convincing performance on semantic segmentation tasks.
Recently, the non-local network \cite{wang2018non} and PSA-Net \cite{zhao2018psanet} employed the self-attention mechanism to capture long-range dependent features by exploiting the correlations among all pixels.
Meanwhile, Fu \textit{et al.} \cite{fu2019dual} proposed a dual attention segmentation network, which creates two fully connected correlation matrices for feature and position attentions, respectively.
This dual attention setting, however, may result in a significant increase of the computational cost.  
To reduce the computational cost of non-local methods, Huang \textit{et al.} \cite{huang2019ccnet} proposed CCNet, which contains an efficient attention module called the criss-cross attention. 
To sum up, most self-attention methods utilize the fully connected correlation matrix to represent the feature correlations.
However, constructing a fully connected correlation matrix is computationally expensive and introduces noise, which may damage the semantic discriminatory power of the features.
In addition, the global information learning in these models remains limited, since only low-level reasoning is performed in the image space.
In our PSGR module, we perform global reasoning in the graph space in an effective way, with particular emphasis on the relation between each uncertain node and the nodes connected to it strongly. 

\subsection{Graph Reasoning for Semantic Segmentation}
Many graph-based methods have been proposed for semantic image segmentation due to their superior relation reasoning capabilities.
Li \textit{et al.} \cite{li2018beyond} developed a novel approach to learning graph representations from 2D feature maps for visual recognition, which uses pixel clustering and feature similarity measurement to transform an image to a graph structure.
Chen \textit{et al.} \cite{chen2019graph} performed relational reasoning by projecting a set of features that are globally aggregated over the coordinate space into an interaction space.
Graph reasoning has also been applied to medical image segmentation.
Soberanis-Mukul \textit{et al.} \cite{soberanis2020uncertainty} combined uncertainty analysis and graph convolutional network (GCN) to refine organ segmentation in CT images.
Liu \textit{et al.} \cite{liu2020cross} utilized a predefined pseudo landmark to project mammogram images to the graph space and then introduced a bipartite GCN to endow DCNN segmentation networks with the cross-view reasoning ability.
However, the feature mapping strategies used in these methods rely highly on either the prior knowledge or a predetermined number of nodes, which tends to result in limited generalizability and adaptiveness.
Hu \textit{et al.} \cite{hu2020class} constructed the graph in a pixel-wise and class-wise manner and performed graph reasoning on dynamically sampled pixels, which avoids all those predetermined and inflexible feature projections and exploits contextual information for semantic segmentation. 
However, the connections are only restricted among those sampled pixels, which may lead to an insufficient aggregation of effective information.
Li \textit{et al.} \cite{li2020spatial} constructed the fully connected graph in a pixel-wise way and organized the graph reasoning as a spatial pyramid. 
However, similar to those self-attention methods, the semantic discriminatory of the features may be ignored when the feature maps are represented by fully connected graphs. 
In contrast, in our PSGR module, we construct a sparse graph from the perspective of the message passing mechanism of GNN, where each node can selectively connect to the nodes from which it can gain more effective information.
This design can facilitate GNN to capture the long-range information in the graph reasoning stage.
\begin{figure*}[tbh]
\centering
\includegraphics[width=1\textwidth]{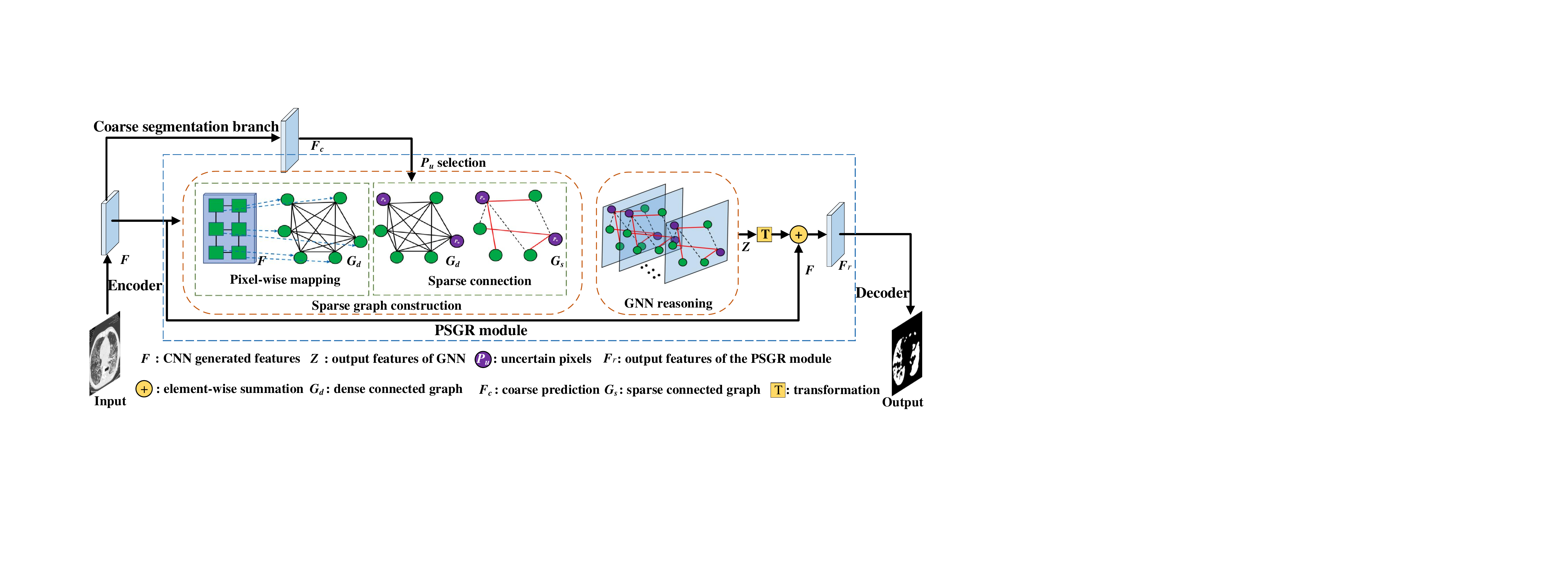}
\caption{Diagram of the proposed segmentation model, including a segmentation backbone, a coarse segmentation branch, and the proposed PSGR module.}
\label{fig1}
\end{figure*}

\section{Preliminaries}
Before introducing the proposed method, we first give the preliminaries of GNN.

An attributed and weighted graph $G$ with $N$ nodes is denoted by $(A, H)$, where $A \in \mathcal{R}^{N \times N}$ is the graph adjacency matrix, $H \in \mathcal{R}^{N \times c}$ is the node feature matrix, and $c$ is the dimensionality of the feature at each node.
The node latent feature matrix $Z$, which represents the embedded node features in the latent space $\mathcal{Z}$, can be formally expressed as follows:
\begin{equation}
    Z^{(k)} = F(A^{(k-1)}, Z^{(k-1)}; \theta^{(k)}),
\end{equation}
where $k$ denotes the $k-th$ layer of GNN, $A^{(k-1)}$ is the graph adjacency matrix computed by the $(k-1)-th$ layer of the GNN, $\theta^{(k)}$ is the ensemble of trainable parameters in the $k-th$ layer, and $F(\cdot)$ is the forward function to aggregate and transform the messages across the nodes. Particularly, $Z^{0}=H$. 
Many previous studies specified different definitions of function $F(\cdot)$ \cite{gilmer2017neural,hamilton2017inductive} such as the graph convolution neural network (GCN) \cite{kipf2016semi} and higher-order GCN (HO-GCN) \cite{morris2019weisfeiler}. The GCN combines the information of the neighborhoods as the node representation linearly. The HO-GCN takes higher-order graph structures into account, which is important to capture the long-range information in the graph.  

\section{Methods}
Our COVID-19 pneumonia segmentation model consists of a segmentation backbone, a coarse segmentation branch, and the proposed PSGR module that is inserted into the segmentation backbone. The diagram of this model is illustrated in Fig. \ref{fig1}. We now delve into its details.
\subsection{PSGR Module}
The PSGR module is composed of two components: sparse graph construction and long-range information reasoning (see Fig. \ref{fig1}). It aims to improve the effectiveness of information gain on uncertainly segmented pixels and hinder the noise propagation in long-range information reasoning, which can further boost the segmentation performance especially on those uncertain pixels.

\subsubsection{Sparse Graph Construction}
Let the feature map generated by the segmentation backbone be denoted by $X \in \mathcal{R}^{h \times w\times c }$ where $h \times w$ is the image size and $c$ is the number of channels. The node feature matrix $H$ can be obtained by reshaping $X$ to the size of $N \times c$, where $N = h \times w$.
After mapping each pixel to a graph node, the constructed graph $G=(A, H)$ preserves the inherent information of each pixel and can provide precise pixel-wise information for global reasoning.
The adjacency matrix $A$ encodes the connection pattern among nodes, which indicates, for each node $v_{i}$, the information is aggregated from which of its neighborhoods. Since it is neither unreasonable nor computationally tractable to fully connect all nodes, we propose a strategy to construct a sparsely-connected graph $G_{s}$ based on the information theory.

\textbf{Connectivity Distribution Matrix.}
Suppose two pixels $p_{i}$ and $p_{j}$ are mapped to two nodes $v_{i}$ and $v_{j}$, respectively. The connectivity between $v_{i}$ and $v_{j}$ is measured by the inner product of the features of $p_{i}$ and $p_{j}$. Thus, a larger connectivity between $v_{i}$ and $v_{j}$ indicates higher similarity between $p_{i}$ and $p_{j}$. The feature similarity matrix $S$ is defined as follows to ensure that its diagonal elements are zero: 
\begin{equation}
    S = H H^{T} - H H^{T} \odot I,
\end{equation}
\noindent where $\odot$ is element-wise product, and $I$ is the identity matrix. The feature similarity matrix $S$ can be regarded as the adjacency matrix of a densely-connected graph $G_{d}$ (see Fig. \ref{fig1}).
Then, we construct the normalized node connectivity distribution matrix $\hat{S}$ by computing the graph Laplacian:
\begin{equation}
    \hat{S} = D^{-\frac{1}{2}} S D^{-\frac{1}{2}},
\end{equation}
where $D$ is the degree matrix of $S$. Note that the $i$-th line in $\hat{S}$, denoted by $\hat{S}_{i:}$, representing the connectivity probability distribution between $v_{i}$ and any other nodes and $\sum \hat{S}_{i:}=1$. 

\textbf{Node Information Score.}
For each node $v_{i}$, we define an information score (IS) to measure the information quantity that $v_{i}$ gains from each of its neighbors, shown as follows:
\begin{equation}
    IS_{i} = \left \| \hat{S}_{i:}^{T} \otimes H \right \|_{\Tilde{L}_{1}},
\end{equation}
where $\left \| \cdot \right \|_{\Tilde{L}_{1}}$ is line-wise $L_{1}$ norm, and $\otimes$ is the scalar-multiplication between each line of two matrices.

\textbf{Sparse Connection Adjacency Matrix.}
The key to construct a sparsely-connected graph is the criterion that can guide edge pruning. We divide all nodes into certain nodes and uncertain nodes and then define the criterion as: (1) the connection between any pair of certain nodes is removed, and (2) for each uncertain node, only the connections between it and $K$ neighbors with highest \textit{IS} values are preserved. 
The certainty of each node is determined based on the predictions made by the coarse segmentation branch. 
Specifically, for each node, we calculate the difference between the largest and second largest predicted probabilities of the corresponding pixel belonging to a region. 
Then, we select $R_{u} \times N$ nodes with lowest probability difference as uncertain nodes, where $R_{u}$ is the uncertain nodes selection ratio.   
As a result, each element of the sparse connection adjacency matrix ($\widetilde A$) can be formally expressed as:
\begin{equation}
    \widetilde{A}_{ij} = \{ \hat{S}_{ij} \mid v_{i} \in \Omega_u, v_{j} \in topK \lceil IS_{i} \rceil \},
\end{equation}
where $\Omega_{u}$ is the set of uncertain nodes, and $topK \lceil IS_{i} \rceil$ generates a set containing $K$ neighbors of $v_{i}$ with highest \textit{IS} values. The hyer-parameter $K$ is empirically set to $N/2$ for this study. Then, we can obtain a sparsely-connected graph $G_{s}=(\widetilde A,H)$.

\subsubsection{Long-Range Information Reasoning with HO-GNN}
The higher-order information, which is aggregated from global neighbors via multi-hops, is difficult to capture but important in reasoning the contextual relations in the graph.
Since HO-GNN \cite{morris2019weisfeiler} is a powerful tool to capture both local and global information in graph-structured data, we utilize HO-GNN to perform graph reasoning in our PSGR module.

The way that HO-GNN aggregates and propagates the information can be formulated as:
\begin{equation}
    Z^{(k)}(v_{i}) = \mathcal{F}(Z^{(k-1)}(v_{i}) \theta^{(k-1)}_{1} +  
    \sum\limits_{v_{j} \in \Phi(v_{i}) }Z^{(k-1)}(v_{j}) \theta^{(k-1)}_{2}),
\end{equation}
where $\Phi(v_{i})=N_{l}(v_{i}) \cup N_{g}(v_{i})$ is the union of the local and global neighborhoods of node $v_{i}$, $Z(v_{i})$ is the latent feature of $v_{i}$, $\mathcal{F}(\cdot)$ is a nonlinear transformation function ($e.g.$, sigmoid), $\theta_{1}$ and $\theta_{2}$ are trainable parameters, and $k$ is the index of layers.
In the stage of graph reasoning, each uncertain node can aggregate information from its local and global neighborhoods, enabling the retrieval of long-range contextual dependencies.

Once obtaining the feature map produced by HO-GNN ($i.e.$, $Z$), we first reshape it back to the size of $h\times w\times c$, and then fuse it with the input feature map $F$ via element-wise summation to generate the output feature map of the PSGR module, denoted by  $F_r$.

\subsection{Segmentation Model with PSGR Module}
\label{seciv-b}
\begin{figure*}[t]
    \centering
    \raggedright
    \centering
    \includegraphics[width=0.9\textwidth]{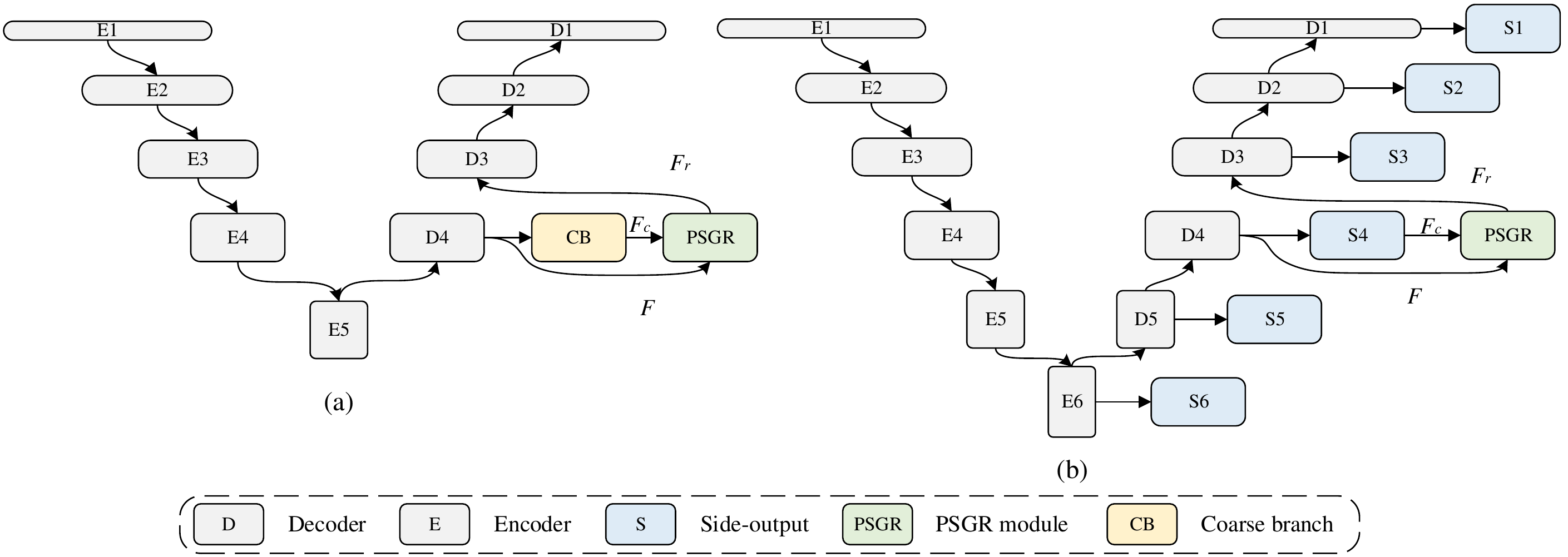}
    \caption{Deploying the coarse segmentation branch and proposed PSGR module in the segmentation backbone. (a) and (b) represent U-Net and U$^2$-Net, respectively. See \ref{seciv-b} for details.}
    \label{fig2}
\end{figure*}

\textbf{Segmentation Backbone.}
For this study, we choose two widely-used baselines as the segmentation backbones, \textit{i.e.}, U-Net \cite{ronneberger2015u} and U$^2$-Net \cite{qin2020u2}. 
The former has shown convincing and robust performance on a large variety of medical image segmentation tasks, and the latter has special two-level nested U-structure which can help to capture abundant contextual information and thereby obtained superior performance on several computer vision tasks.
Here we adopt all default configurations used in the official implementations\footnote{\url{https://github.com/milesial/Pytorch-UNet}},\footnote{\url{https://github.com/xuebinqin/U-2-Net}}, except for replacing the transposed convolution with the bi-linear interpolation in U-Net.

\textbf{Coarse Segmentation Branch.}
To determine uncertain nodes, we need a coarse segmentation branch to predict the rough probability of a pixel belonging to each region.
Since the features in deep stages may have too low a spatial resolution to recover the details, we place the coarse segmentation branch after the fourth stage of the decoder in the backbone (see Fig. \ref{fig2}), where the feature map has 1/8 size of the input image.
For U-Net, we first apply a layer sequence of $3\times3  Conv+BN+ReLU+1\times1 Conv$ to produce the coarse prediction map \textbf{$F_c$} where the middle channel is set to 128.
Besides feeding \textbf{$F_c$} to the PSGR module, we also upsample it to the input size as an auxiliary deep supervision.
Considering U$^2$-Net has side-output for each stage of the decoder, we directly adopt its fourth side-output as \textbf{$F_c$}.

\textbf{Deploying PSGR Module.}
As illustrated in Fig. \ref{fig1} and Fig. \ref{fig2}, the PSGR module takes both \textbf{$F$} and \textbf{$F_c$} as inputs, and directly produces refined feature map \textbf{$F_r$}, which is enhanced with global long-range dependencies.
Inside our PSGR module, we specially apply $1\times1$ convolutions to keep the size and channel number of \textbf{$F$} and \textbf{$F_r$} consistent.
Due to its pixel-wise mapping strategy and flexible adaptability, our PSGR module can also be easily incorporated into any other segmentation networks in an end-to-end-training fashion.

\textbf{Loss Function and Supervision Manner.}
Since we adopt the coarse segmentation result $s_{coarse}$ for auxiliary supervision, the loss function is defined as follows:
\begin{equation}
    L =L_{seg}(s_{main}, y) + \lambda L_{seg}(s_{coarse}, y),
\label{eq2}
\end{equation}
where $s_{main}$ is the segmentation results produced by the backbone, $y$ is the ground truth, and the weighting parameter $\lambda$ is set to 0.5 for all experiments without further tuning. Each segmentation loss $L_{seg}$ is the sum of the binary cross-entropy (BCE) loss and Dice loss, shown as follows:
\begin{equation}
    L_{seg} =\ell_{BCE} + \ell_{Dice}.
\label{eq1}
\end{equation}

\begin{table}[htp]
\caption{Details of three public datasets.}
\setlength\tabcolsep{10pt}
\begin{tabular}{l|c|c}
\hline
Datasets         & Slice number  & Resolution \\ \hline
COVID19-CT-100 \cite{covid2020dataset}           & 100       & 512$\times$512             \\
COVID19-CT-Seg20 \cite{Ma2020Covid}     & 1844      & 512$\times$512-630$\times$630  \\
MosMedData \cite{morozov2020mosmeddata} & 785       & 512$\times$512             \\ \hline
\end{tabular}
\label{tab1}
\end{table}

\begin{table*}[t]
\centering
\setlength\tabcolsep{5pt}
\caption{Quantitative results of different methods on three public datasets. The best and second best results are shown in \textcolor{red}{red} and \textcolor{blue}{blue}, respectively. The values of mIoU, SEN, SPE, and DSC are in percentage terms and the value of HD is in $mm$.}
\begin{tabular}{l|ccccc|ccccc|ccccc}
\hline
\multicolumn{1}{c|}{}                          & \multicolumn{5}{c|}{COVID19-CT-100}  & \multicolumn{5}{c|}{COVID19-CT-Seg20} & \multicolumn{5}{c}{MosMedData} \\ \cline{2-16} 
\multicolumn{1}{c|}{\multirow{-2}{*}{Methods}} & mIoU   & SEN   & SPE   & DSC & HD  & mIoU  & SEN   & SPE   & DSC   & HD   & mIoU  & SEN  & SPE  & DSC  & HD \\ \hline
FCN-8s \cite{long2015fully}                    &71.85   &66.47  &93.56  &58.11  &104.68 &82.54  &84.10  &98.02  &73.60  &51.47  &70.51  &\textcolor{blue}{80.75}  &97.08  &53.33  &84.43 \\
DeepLabv3+ \cite{chen2018encoder}              &79.45   &79.58  &97.55  &71.70&93.09&81.26  &81.61  &95.35  &42.79     &182.14&74.14  &74.65 &97.26 &57.16 &102.78 \\
U-Net++ \cite{zhou2018unet++}                  &77.64   &77.26  &97.28  &69.04&91.73&80.73  &79.61  &96.75  &70.34  &63.01 &73.39  &75.67 &96.13 &59.08 &88.21  \\
Attention U-Net \cite{oktay2018attention}      &77.71   &74.75  &97.56  &68.93&92.15&80.70  &82.92  &97.41  &71.27  &64.91 &74.62  &\textcolor{red}{81.32} &97.63 &59.34 &95.16  \\
DANet \cite{fu2019dual}                        &73.57   &66.30  &92.76  &61.34&99.11&81.59  &\textcolor{red}{88.78}  &99.13  &73.82  &114.69 &73.47  &75.00 &95.80 &56.07 &74.04  \\ 
CCNet \cite{huang2019ccnet}                    &75.24   &69.55  &95.92  &63.99&98.03&81.27  &\textcolor{blue}{86.61}  &99.16  &73.93  &90.84 &72.02  &79.16 &96.29 &54.83 &83.07  \\ 
Inf-Net \cite{fan2020inf}                      &81.62   &76.50  &98.32  &74.44&86.81&64.62  &69.46  &99.02  &63.38  &79.68 &74.32  &62.93 &93.45 &56.39 &71.77  \\ 
MiniSeg \cite{qiu2021miniseg}                  &82.15   &\textcolor{blue}{84.95}  &97.72  &75.91&74.42&84.49  &85.06  &99.05  &76.27  &51.06 &78.33  &79.62 &97.71 &64.84 &71.69  \\ \hline
\textbf{U-Net+PSGR}                            &\textcolor{blue}{86.58}    &83.62   &\textcolor{blue}{98.86}   &\textcolor{blue}{83.16}&\textcolor{blue}{50.68} &\textcolor{red}{87.88}   &77.83   &\textcolor{red}{99.78}   &\textcolor{red}{78.58}  &\textcolor{blue}{46.86}  &\textcolor{blue}{80.16}  &72.73 &\textcolor{blue}{99.86} & \textcolor{blue}{66.95}    &\textcolor{blue}{68.49}   \\
\textbf{U$^2$-Net+PSGR}                       &\textcolor{red}{87.92}  &\textcolor{red}{85.89}  &\textcolor{red}{98.95}  &\textcolor{red}{84.83}&\textcolor{red}{42.85} &\textcolor{blue}{87.55} &79.78  &\textcolor{blue}{99.75} &\textcolor{blue}{78.32}  &\textcolor{red}{43.51}   &\textcolor{red}{80.52} &72.30  &\textcolor{red}{99.88}    &\textcolor{red}{67.27} &\textcolor{red}{64.45}\\ \hline
\end{tabular}
\label{compare-results1}
\end{table*}

\section{Experimental Setup}
\subsection{Datasets}
For this study, three public COVID-19 pneumonia CT segmentation datasets, \textit{i.e.}, COVID-19 CT Segmentation Dataset (COVID19-CT-100) \cite{covid2020dataset}, COVID-19 CT Lung and Infection Segmentation Dataset (COVID19-CT-Seg20) \cite{Ma2020Covid}, and MosMedData \cite{morozov2020mosmeddata} were used to evaluate our method.
The COVID19-CT-100 dataset was collected by the Italian Society of Medical and Interventional Radiology\footnote{\url{https://sirm.org/category/senza-categoria/covid-19/}} which consists of 100 COVID-19 infected CT slices from \textgreater 40 patients.
Since the annotations of different infected regions (ground-glass opacity (GGO) and consolidation) were provided, we follow \cite{qiu2021miniseg} and \cite{fan2020inf} to evaluate the segmentation performance of our method on bi-class segmentation and multi-class segmentation, respectively.
The COVID19-CT-Seg20 dataset contains 20 COVID-19 CT images where lungs and infections were annotated by two radiologists and verified by an experienced radiologist. Here we only focused on the segmentation of the COVID-19 infection, since it is more challenging and important.
The MosMedData dataset was collected by the Research and Practical Clinical Center for Diagnostics and Telemedicine Technologies of the Moscow Health Care Department. 
A total of 50 CT scans, each having less than 25\% lung infected, were selected and manually labeled by experts.
Considering the limited scans and large inter-slice spacing of those volumetric data, we followed previous work \cite{fan2020inf, qiu2021miniseg} to perform 2D segmentation on all datasets.
As a result, we totally have 100, 1844, and 785 2D CT slices from COVID19-CT-100, COVID19-CT-Seg20, and MosMedData, respectively. 
The details of these datasets were shown in Table \ref{tab1}.

\subsection{Implementation Details}
In the pre-processing step, we simply normalized the intensities of each slice to zero mean and unit variance.
During the training phase, we first applied data augmentation techniques on the fly to reduce potential overfitting, including random scaling ($0.8$ to $1.2$), random rotation ($\pm 15^{\circ}$), random intensity shift of ($\pm0.1$) and intensity scaling of ($0.9$ to $1.1$). 
Then, we cropped or padded each image to a size of $512\times512$.
The training iterations were set to 200 epochs with a linear warmup of the first 5 epochs. We trained the model using the Adam optimizer with a batch size of 8 and synchronized batch normalization. 
The initial learning rate was set to $1e^{-3}$ and decayed by $(1-\frac{current\_epoch}{max\_epoch})^{0.9}$.
We also regularized the training with an $l_2$ weight decay of $1e^{-5}$. 
The uncertain pixel selection ratio $R_{u}$ was set to 0.005, 0.01, and 0.005 on COVID19-CT-100, COVID19-CT-Seg20, and MosMedData, respectively.

We used five-fold cross-validations for bi-class segmentation where the data division in \cite{qiu2021miniseg}\footnote{\url{https://github.com/yun-liu/MiniSeg}} was adopted.
As for multi-class segmentation, we adopted the data division in \cite{fan2020inf}\footnote{\url{https://github.com/DengPingFan/Inf-Net}} and divided the COVID19-CT-100 dataset into a training set, a validation set, and a test set.
All experiments were conducted on a workstation with 2 NVIDIA TITAN RTX GPUs.

\subsection{Baselines and Evaluation Metrics}
Besides evaluating the effectiveness of our PSGR module against U-Net and U$^2$-Net in the ablation study, we also compared our approach with eight segmentation baselines, $i.e.$, FCN-8s \cite{long2015fully}, DeepLabv3+ \cite{chen2018encoder}, U-Net++ \cite{zhou2018unet++}, Attention U-Net \cite{oktay2018attention}, Inf-Net \cite{fan2020inf}, MiniSeg \cite{qiu2021miniseg}, DANet \cite{fu2019dual}, and CCNet \cite{huang2019ccnet}.
U-Net++ and Attention U-Net are two well-performing baselines in medical image segmentation, while FCN-8s and DeepLabv3+ are two popular baselines in semantic segmentation.
MiniSeg and Inf-Net are two state-of-the-art (SOTA) models which have shown convincing performance in COVID-19 segmentation.
DANet and CCNet are introduced as two cutting-edge attention-based networks which also focus on enhancing long-range dependencies for semantic segmentation models.

We adopted six metrics to assess the performance of segmentation models, including the mean intersection over union (mIoU), Dice similarity coefficient (DSC), sensitivity (SEN), specificity (SPE), Hausdorrf distance (HD), and mean absolute error (MAE).
Specifically, mIoU, DSC, SEN, and SPE are four overlap-based metrics, each ranging from 0 to 1 and a larger value indicating better performance.
HD is a shape distance-based metric, which can be used to measure the dissimilarity between the surfaces / boundaries of the segmentation result and the ground-truth.
MAE can represent the dissimilarity between the segmentation result and the ground-truth.
As for HD and MAE, a lower value indicates a better segmentation result.
We followed \cite{qiu2021miniseg} to adopt mIoU, SEN, SPE, DSC, and HD for bi-class infection segmentation tasks, and followed \cite{fan2020inf} to choose DSC, SEN, SPE, and MAE for the multi-class infection segmentation task.

\begin{figure*}[t]
    \centering
    \raggedright
    \centering
    \includegraphics[width=\textwidth]{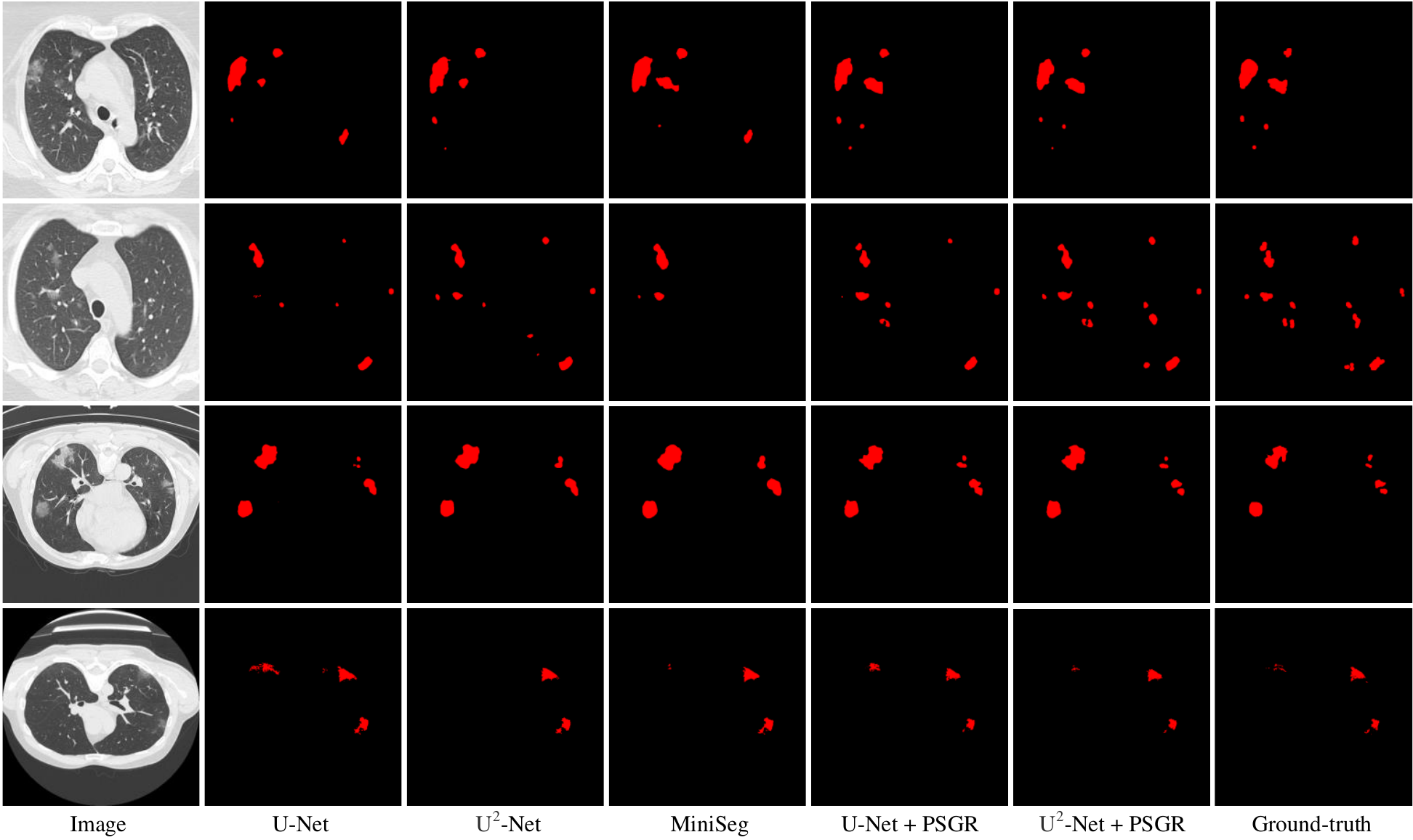}
    \caption{Visualization of the bi-class infection segmentation results produced by our models and three competing ones on the COVID19-CT-100 (row 1 and 2), COVID19-CT-Seg20 (row 3), and MosMedData (row 4) datasets. Comparing the results in column 2 and 5 (or column 3 and 6), we can conclude that the segmentation improvements should be attributed to the strong long-range information reasoning ability of our PSGR module.}
    \label{binary-class}
\end{figure*}
\begin{table*}[tbh]
\centering
\caption{Quantitative results of different methods for multi-class infection segmentation on the COVID119-CT-100 dataset. The best and second best results are shown in \textcolor{red}{red} and \textcolor{blue}{blue}, respectively. The values of DSC, SEN, SPE, and MAE are in percentage terms. $*$ represents using extra training data.}
\begin{tabular}{l|cccc|cccc|cccc}
\hline
\multirow{2}{*}{Methods} & \multicolumn{4}{c|}{GGO}  & \multicolumn{4}{c|}{Consolidation} & \multicolumn{4}{c}{Average} \\ \cline{2-13} 
& DSC   &SEN    &SPE    &MAE    &DSC    &SEN    &SPE    &MAE    &DSC    &SEN    &SPE    &MAE    \\ \hline
DeepLabv3+ \cite{chen2018encoder}      &44.3   &\textcolor{blue}{71.3}   &82.3   &15.6   &23.8   &31.0   &70.8   &7.7   &34.1   &51.2   &76.6   &11.7    \\
FCN-8s \cite{long2015fully}            &47.1   &53.7   &90.5   &10.1   &27.9   &26.8   &71.6   &5.0   &37.5   &40.3   &81.1   &7.6    \\
U-Net \cite{ronneberger2015u}          &44.1   &34.3   &\textcolor{blue}{98.4}   &8.2   &40.3   &41.4   &96.7   &5.5   &42.2   &37.9   &97.6   &6.6    \\
Inf-Net (FCN-8s)$^*$ \cite{fan2020inf}     &\textcolor{red}{64.6}   &\textcolor{red}{72.0}   &94.1   &7.1   &30.1   &23.5   &80.8   &4.5   &47.4   &47.8   &87.5   &5.8    \\
Inf-Net (U-Net)$^*$ \cite{fan2020inf}      &\textcolor{blue}{62.4}   &61.8   &96.6   &6.7   &45.8   &\textcolor{blue}{50.9}   &96.7   &4.7   &54.1   &\textcolor{blue}{56.4}   &96.7   &5.7    \\ \hline
\textbf{U-Net+PSGR}                    &62.3   &69.3   &97.9   &\textcolor{blue}{3.2}   &\textcolor{blue}{49.0}   &\textcolor{red}{65.3}   &\textcolor{blue}{98.4}   &\textcolor{blue}{2.1}    &\textcolor{red}{55.7}    &\textcolor{red}{67.3}  &\textcolor{blue}{98.2}  &\textcolor{blue}{2.7} \\
\textbf{U$^2$-Net+PSGR}                &60.2   &58.5   &\textcolor{red}{98.9}   &\textcolor{red}{2.9}   &\textcolor{red}{49.8}   &50.1   &\textcolor{red}{99.2}   &\textcolor{red}{1.6}    &\textcolor{blue}{55.0}   &54.3   &\textcolor{red}{99.1}   &\textcolor{red}{2.3}    \\ \hline
\end{tabular}
\label{compare-results2}
\end{table*}

\section{Results and Discussions}
\begin{table}[tb]
\centering
\setlength\tabcolsep{5pt}
\caption{Ablation studies of our proposed PSGR module on the COVID19-CT-100 datasets. The best results are shown in \textcolor{red}{red}. The values of mIoU, SEN, SPE, and DSC are in percentage terms and the value of HD is in $mm$.}
\begin{tabular}{l|ccccc}
\hline
\multirow{2}{*}{Methods} & \multicolumn{5}{c}{COVID19-CT-100} \\ \cline{2-6} 
& mIoU      & SEN       & SPE       & DSC       & HD    \\ \hline
(a) U-Net \cite{ronneberger2015u}     &77.56  &72.24  &97.71  &68.37  &94.25  \\
(b) U-Net+CSB                          &82.01  &76.81  &98.58  &76.88  &71.01  \\
(c) U-Net+CSB+PSGR         &\textcolor{red}{86.58}    &\textcolor{red}{83.62}   &\textcolor{red}{98.86}   &\textcolor{red}{83.16}&\textcolor{red}{50.68}    \\ \hline\hline
(d) U$^2$-Net \cite{qin2020u2}        &80.46  &76.92  &97.62  &75.87  &75.87  \\
(e) U$^2$-Net+PSGR        &\textcolor{red}{87.92}  &\textcolor{red}{85.89}  &\textcolor{red}{98.95}  &\textcolor{red}{84.83}&\textcolor{red}{42.85}      \\ \hline
\end{tabular}
\label{ablation}
\end{table}
\subsection{Comparative Experiments}
\label{sub-ce}
\textbf{Performance in Bi-class Infection Segmentation.}
Table \ref{compare-results1} gives the performance of our models and eight competing ones, including FCN-8s \cite{long2015fully}, DeepLabv3+ \cite{chen2018encoder}, U-Net++ \cite{zhou2018unet++}, Attention U-Net \cite{oktay2018attention}, DANet \cite{fu2019dual}, CCNet \cite{huang2019ccnet}, Inf-Net \cite{fan2020inf}, and MiniSeg \cite{qiu2021miniseg} in bi-class infection segmentation on the COVID19-CT-100, COVID19-CT-Seg20, and MosMedData datasets.
It shows that our models (\textit{i.e.}, U-Net equipped with our PSGR module (U-Net+PSGR) and U$^{2}$-Net equipped with our PSGR module (U$^{2}$-Net+PSGR)) outperform all competing methods substantially and consistently in terms of DSC and mIoU, indicating that the segmentation results of our models match well with the ground-truth.
Across all metrics, U$^{2}$-Net+PSGR and U-Net+PSGR achieve the overall best and second best performance, respectively.
Meanwhile, comparing to two self-attention based SOTAs, \textit{i.e.}, DANet \cite{fu2019dual} and CCNet \cite{huang2019ccnet}, our models achieve clearly superior segmentation results, which tend to show the strong ability in capturing long-range dependencies for COVID-19 infection segmentation.
At last, it is remarkable that using our PSGR module can significantly reduce the HD values when comparing to any competing models, which demonstrates that the boundaries detected in our segmentation results match the ground-truth boundaries very well.

\begin{figure*}[t]
    \centering
    \raggedright
    \centering
    \includegraphics[width=\textwidth]{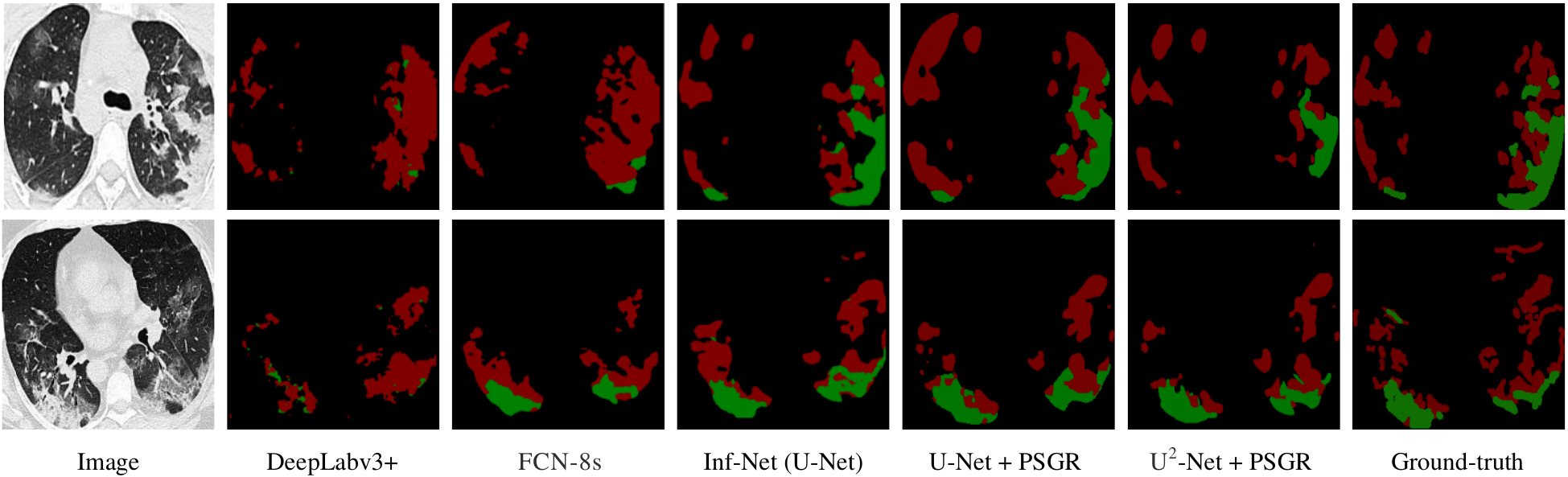}
    \caption{Visualization of the multi-class infection segmentation results on the COVID19-CT-100 dataset produced by our models and three competing ones. The regions of GGO and Consolidation are highlighted in \textcolor{red}{red} and \textcolor{green}{green}, respectively.}
    \label{multi-class}
\end{figure*}
\begin{figure*}[t]
    \centering
    \raggedright
    \centering
    \includegraphics[width=\textwidth]{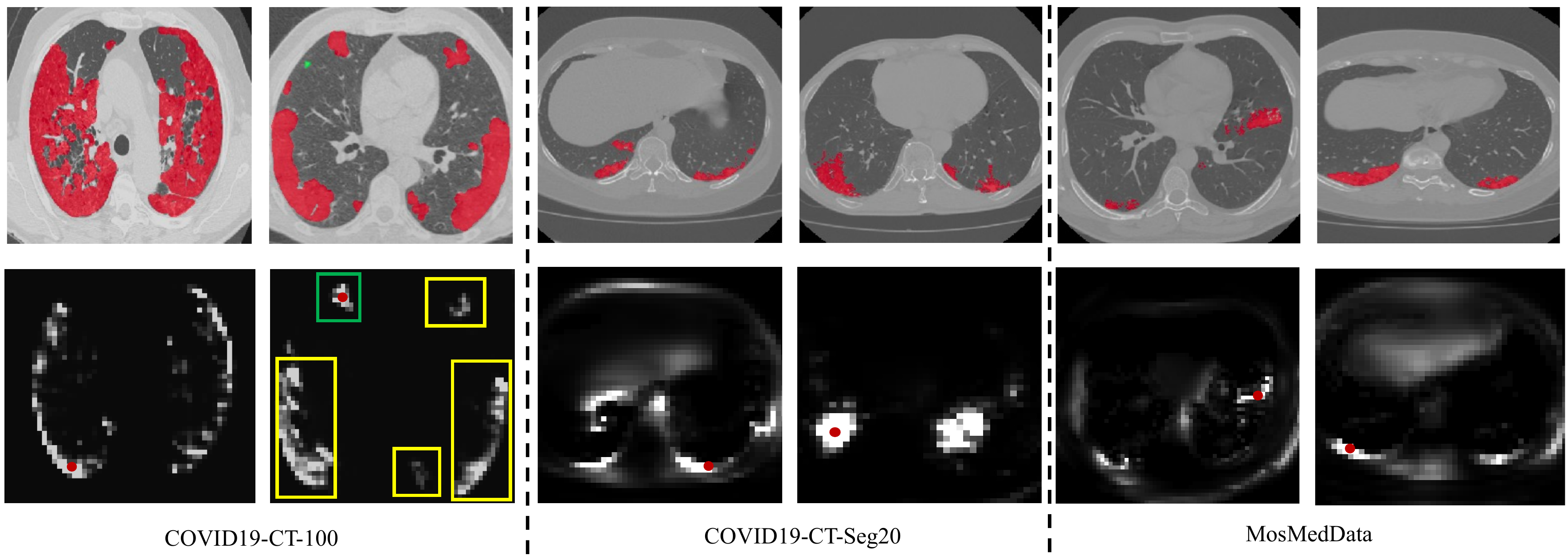}
    \caption{Visualization of the ability of our PSGR module to capture long-range dependencies on three datasets.
    Given a pixel (\textcolor{red}{red} dot) in an infectious region (\textcolor{green}{green} box), our PSGR module can highlight other foreground pixels (\textcolor{yellow}{yellow} boxes) on the entire image, where the contextual information exists.}
    \label{attention}
\end{figure*}
\textbf{Performance in Multi-class Infection Segmentation.}
Table \ref{compare-results2} gives the performance of our models and five competing ones, including FCN-8s \cite{long2015fully}, U-Net \cite{ronneberger2015u}, DeepLabv3+ \cite{chen2018encoder}, and Inf-Net \cite{fan2020inf} (with two backbones), in multi-class infection segmentation on the COVID19-CT-100 dataset.
It reveals that Inf-Net (FCN-8s) \cite{fan2020inf} has better segmentation performance than DeepLabv3+ \cite{chen2018encoder}, FCN-8s \cite{long2015fully}, and U-net \cite{ronneberger2015u}, and achieve best DSC and SEN on the GGO segmentation task. Our U-Net+PSGR and U$^2$-Net+PSGR achieve best performance across all metrics in the segmentation of consolidation, which is more challenging since each consolidation region tends to have a tiny size.
In addition, our models have consistently and significantly lower MAE than other models on both GGO and consolidation segmentation, which indicates again that our segmentation results have less mismatched predictions.
In summary, both U-Net+PSGR and U$^2$-Net+PSGR achieve overall best performance across all metrics.
It is worth noting that, different from Inf-Net, which actually performs semi-supervised segmentation using 1600 extra unlabelled CT slices, we only use 50 CT slices from the COVID19-CT-100 dataset to train our model for this challenging segmentation task.

All these convincing results on three datasets for both bi-class and multi-class segmentation tasks demonstrate the effectiveness and strong generalizability of our U-Net+PSGR and U$^2$+PSGR models.
\begin{figure*}[t]
    \centering
    \raggedright
    \centering
    \includegraphics[width=\textwidth]{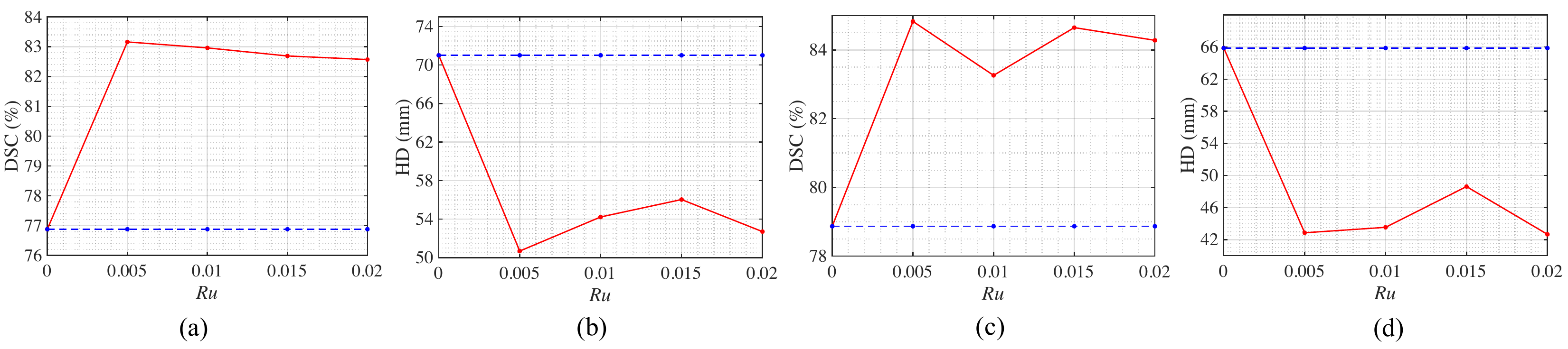}
    \caption{Impact of $R_{u}$ on segmentation performance: (a) DSC of U-Net+PSGR v.s. $R_{u}$, (b) HD of U-Net+PSGR v.s. $R_{u}$, (c) DSC of U$^2$-Net+PSGR v.s. $R_{u}$, and (d) HD of U$^2$-Net+PSGR v.s. $R_{u}$. The \textcolor{blue}{blue} dashlines are the baseline results where the proposed PSGR module is not integrated. Higher DSC values or lower HD values indicate better segmentation performance.}
    \label{np-analysis}
\end{figure*}
\subsection{Ablation Study}
\label{sub-as}
We conducted an ablation study on the COVID19-CT-100 dataset under a bi-class segmentation setting to evaluate the effectiveness of our PSGR module.
We compared our U-Net+PSGR with its baseline U-Net \cite{ronneberger2015u} and U$^2$-Ne+PSGR to U$^2$-Net \cite{qin2020u2}. Besides, since U-Net has no deep supervision structure, we added a coarse segmentation branch (CSB) to U-Net to provide deep supervision and reported the results, too.
The results in Table \ref{ablation} show that
(1) solely introducing CSB to U-Net can improve DSC from $68.37\%$ to $76.88\%$;
(2) integrating our PSGR module to U-Net or U$^2$-Net can substantially improve the segmentation performance in terms of all metrics; and
(3) U$^2$-Net+PSGR achieves the best performance with DSC of $84.83\%$ and HD of $42.85\%$.
The consistent performance gains over baselines demonstrate the effectiveness of our PSGR module for COVID-19 infection segmentation.

\subsection{Visualization of Segmentation Results}
Four COVID-19 CT images from three datasets, the bi-class infection segmentation results produced by U-Net, U$^2$-Net, MiniSeg, and our models, and the corresponding ground-truths are visualized in Fig. \ref{binary-class}.
It shows that, compared to three competing models, our U-Net+PSGR and U$^2$-Net+PSGR can generate the infectious regions that match better with the ground-truths, especially when those regions are disperse and tiny.
Comparing the results of U-Net+PSGR and U-Net (or U$^2$-Net+PSGR and U$^2$-Net), we can conclude that the improvements of segmentation performance should be attributed to the strong long-range information reasoning ability of our PSGR module.
To further demonstrate the ability of our PSGR module to capture long-range dependencies, we chose six images (two from each dataset) as a case study, randomly selected a foreground pixel on each image, and visualized the corresponding row in the sparse connection adjacency matrix ($\widetilde{A}$) in Fig. \ref{attention}. 
It reveals that our PSGR module can accurately capture long-range dependencies with respect to specific semantic information. For instance, the infectious region in the green box is quite difficult to segment since it is tiny and isolated (see Fig. \ref{attention}). Fortunately, given a pixel in this region, our PSGR module can successfully highlight other foreground pixels (highlighted with yellow boxes) from the global, where the useful contextual information exists, to facilitate the segmentation task. 

We also visualized two COVID-19 CT images from the COVID19-CT-100 dataset and the corresponding multi-class infection segmentation results and ground-truths in Fig. \ref{multi-class}. 
It reveals that the results produced our U-Net+PSGR and U$^2$-Ne+PSGR are much more similar to the ground-truths than those generated by DeepLabv3+ and FCN-8s, and are comparable to the results of Inf-Net(U-Net). Note that Inf-Net was trained with a huge number of external data. 

\subsection{Impact of Uncertain Pixels Selection Ratio}
In the proposed PSGR module, the hyperparameter $R_{u}$ represents the ratio of how many uncertain nodes should be selected.
To investigate its impact on the segmentation performance, we plotted the DSC and HD values obtained on the COVID19-CT-100 dataset versus the values of $R_{u}$ in Fig. \ref{np-analysis}.
Since infectious regions occupy around 1\% area on most COVID-19 CT slices, we increased the value of $R_{u}$ from 0 to 0.02 with a step of 0.005.
It shows that, with the increase of $R_{u}$, the segmentation performance of U-Net+PSGR and U$^2$-Net+PSGR tends to first incline and then decline.
It indicates that a large $R_{u}$ degrades the segmentation performance, which may be attributed to the redundant information and noise introduced by excessive uncertain nodes during the graph reasoning process. 
The best value of $R_{u}$, which leads to the highest DSC and lowest HD, is 0.005 for both models.
In addition, Fig. \ref{np-analysis} also indicates that using our PSGR module with different $R_{u}$ values consistently outperforms the baselines (blue dash-lines), which again justifies the robustness and effectiveness of the proposed PSGR module.

\section{Conclusion}
In this paper, we propose an effective graph reasoning module called PSGR to capture long-range contextual information and incorporate it into different segmentation backbones to improve the segmentation of COVID-19 infection in CT images.
The PSGR module has two advantages over existing graph reasoning techniques for semantic segmentation. 
First, the pixel-wise mapping strategy used for mapping an image to a graph not only avoids imprecise pixel-to-node projections but also preserves the inherent information of each pixel. 
Second, the edge pruning method used to construct a sparsely-connected graph results in effective information retrieval and reduces the noise propagation in GNN-based graph reasoning.
Our results show that the segmentation networks equipped with our PSGR module outperform several widely-used segmentation models on three public datasets.
In the future, we plan to extend the proposed solution to contrastive learning settings and thus provide a highly effective pre-training method for various downstream medical image segmentation tasks.

\section*{acknowledgment}
\label{sec:acknowledgment}
We appreciate the efforts devoted by Italian Society of Medical and Interventional Radiology,  Research and Practical Clinical Center for Diagnostics, and Tele-medicine Technologies of the Moscow Health Care Department to collect and share the data for comparing the segmentation algorithms for COVID-19 infection in CT images.  

\ifCLASSOPTIONcaptionsoff
  \newpage
\fi



%
\bibliographystyle{IEEEtran}
\end{document}